\documentclass[pdflatex,sn-mathphys-num]{sn-jnl}

\usepackage{graphicx}%
\usepackage{multirow}%
\usepackage{amsmath,amssymb,amsfonts}%
\usepackage{amsthm}%
\usepackage{mathrsfs}%
\usepackage[title]{appendix}%
\usepackage{xcolor}%
\usepackage{textcomp}%
\usepackage{manyfoot}%
\usepackage{booktabs}%
\usepackage[numbers]{natbib}
\usepackage{algorithm}%
\usepackage{algorithmicx}%
\usepackage{algpseudocode}%
\usepackage{listings}%
\usepackage{cleveref}

\usepackage[final]{changes} 

\definechangesauthor[name={adder}, color=blue]{+} 
\definechangesauthor[name={deleter}, color=red]{-} 
\newcommand{\ADD}[1]{\added[id={+}]{#1}}
\newcommand{\DEL}[1]{\deleted[id={-}]{#1}}

\theoremstyle{thmstyleone}%
\theoremstyle{thmstyletwo}%

\theoremstyle{thmstylethree}%

\begin{document}

\title[Article Title]{Markerless Augmented Reality Registration for Surgical Guidance: A Multi-Anatomy Clinical Accuracy Study}


\author[1,2]{\fnm{Yue} \sur{Yang}}\email{yueyang1@stanford.edu}
\author[1]{\fnm{Fabian} \sur{Necker}}\email{necker@stanford.edu}
\author[1]{\fnm{Christoph} \sur{Leuze}}\email{cleuze@stanford.edu}
\author[1]{\fnm{Michelle} \sur{Chen}}\email{michelle.chen@stanford.edu}
\author[1]{\fnm{Andrey} \sur{Finegersh}}\email{afineger@stanford.edu}
\author[1]{\fnm{Jake} \sur{Lee}}\email{jakejlee@stanford.edu}
\author[1]{\fnm{Vasu} \sur{Divi}}\email{vdivi@stanford.edu}
\author[1]{\fnm{Bruce} \sur{Daniel}}\email{bdaniel@stanford.edu}
\author[1]{\fnm{Brian} \sur{Hargreaves}}\email{bah@stanford.edu}
\author[2]{\fnm{Jie Ying} \sur{Wu}}\email{jieying.wu@vanderbilt.edu}
\author*[1]{\fnm{Fred M} \sur{Baik}}\email{fbaik@stanford.edu}

\affil[1]{\orgdiv{School of Medicine}, \orgname{Stanford University}, \orgaddress{\street{Stanford}, \postcode{94305}, \state{CA}, \country{USA}}}

\affil[2]{\orgname{Vanderbilt Institute of Surgery and Engineering}, \orgaddress{\street{Nashville}, \postcode{37232}, \state{TN}, \country{USA}}}


\abstract{\textbf{Purpose:} In this paper, we develop and clinically evaluate a depth-only, markerless augmented reality (AR) registration pipeline on a head-mounted display, and assess accuracy across small, or low-curvature anatomies in real-life operative settings. \textbf{Methods:} On HoloLens~2, we align Articulated HAnd
Tracking (AHAT) depth to Computed Tomography (CT)-derived skin meshes via (i) depth-bias correction, (ii) brief human-in-the-loop initialization, (iii) global and local registration. We validated the surface-tracing error metric by comparing ``skin-to-bone'' relative distances to CT ground truth on leg and foot models, using an AR tracked tool. We then performed seven intraoperative target trials (feet$\times$2, ear$\times$3, leg$\times$2) during the initial stage of fibula free-flap harvest and mandibular reconstruction surgery, and collected 500+ data per trial. \textbf{Results:} Preclinical validation showed tight agreement between AR-traced and CT distances (leg: median $|\Delta d|$ 0.78\,mm, RMSE 0.97\,mm; feet: 0.80\,mm, 1.20\,mm). Clinically, per-point error had a median 3.9\,mm. Median errors by anatomy were 3.2\,mm (feet), 4.3\,mm (ear), and 5.3\,mm (lower leg), with 5\,mm coverage 92--95\%, 84--90\%, and 72--86\%, respectively. Feet vs.\ lower leg differed significantly ($\Delta$median $\approx$ 1.1\,mm; $p<0.001$). \textbf{Conclusion:} A depth-only, markerless AR pipeline on HMDs achieved $\sim$3--4\,mm median error across feet, ear, and lower leg in live surgical settings without fiducials, approaching typical clinical error thresholds for moderate-risk tasks. Human-guided initialization plus global-to-local registration enabled accurate alignment on small or low-curvature targets, improving the clinical readiness of markerless AR guidance.}




\maketitle

\maketitle

\section{Introduction \& Related Work}\label{sec1}

Augmented reality (AR) head-mounted displays can overlay critical anatomical information directly into the surgeon’s view, improving intraoperative situational awareness \cite{malhotra2023augmented, birlo2022utility}. The Microsoft HoloLens 2 (HoloLens), in particular, has been adopted across procedures ranging from rigid anatomy (spine, cranium) to more mobile targets (e.g., knee) \cite{kerkhof2025depth}. AR guidance is especially attractive for orthopedics and craniofacial surgery, where rigid bony anatomy enables accurate alignment of pre-operative Computed Tomography (CT)/Magnetic Resonance Imaging (MRI) derived models and delivers “X-ray vision” in situ \cite{canton2024feasibility}.

\noindent\textbf{Marker-based AR registration.}
Conventional surgical AR often relies on fiducial markers and retro-reflective spheres, optical tags, or QR/ArUco codes. They are often rigidly attached to the patient or instruments and tracked by cameras \cite{perez2021effect}. These optical tracking systems can achieve sub-millimeter accuracy with high update rates \cite{martin2023sttar} and are common in orthopedic/neurosurgical navigation and robotic workflows when a rigid frame is fixed to bone \cite{condino2021evaluation, siddiqi2021clinical}. However, marker placement and calibration add workflow overhead; rigid fixation reduces adaptability for moving structures; and line-of-sight occlusions or marker displacement can degrade tracking and overlay reliability \cite{furuse2023influence, trinh2022practical}. Moreover, marker-based systems need markers fixed during pre-op imaging of patient so they appear in the scan, and those markers must also remain fixed from imaging through surgery, complicating preparation \cite{duan2025localization}.

\noindent\textbf{Markerless AR registration.}
To streamline workflow and reduce invasiveness, markerless methods use the patient’s own anatomy for alignment. Early vision-based systems matched tooth or bone surfaces from stereo/intra-oral/3D scanners to CT and reported sub-millimetre target registration errors \cite{labadie2004submillimetric, suenaga2015vision}. Subsequent work applied Iterative Closest Points (ICP) alignment to facial/skull surfaces in AR or tablet systems \cite{tu2023exploring, rusinkiewicz2001efficient}. More recently, markerless pipelines have been embedded in HMDs: e.g., ARCUS on HoloLens 2 demonstrated depth-based registration in under 30\,s in a feasibility study but reported errors on the order of 10\,mm, above common clinical thresholds of $<\!5$\,mm \cite{groenenberg2024feasibility}. Other contemporaneous efforts include maxillofacial guidance without physical cutting guides \cite{ury2025markerless} and Artificial Intelligence (AI)-assisted markerless AR with median errors near $\sim$1.4\,mm in controlled setups \cite{shankar2025ai}. Despite progress, many HMD-based, markerless systems still underperform relative to clinical expectations and lack patient-on-table evaluation \cite{qian2017comparison, carbone2025targeting, kerkhof2025depth}.

\noindent\textbf{Depth-based point cloud registration.}
Our work falls under depth-based markerless registration, aligning intraoperative 3D point clouds from Time-of-Flight (ToF) depth sensors to pre-operative CT/MRI models. These methods use geometric cues and are well-suited to modern HMDs, combining global alignment with local refinement. Operating Room (OR) deployment remains challenging because depth sensors exhibit systematic biases on skin \cite{li2024evd}. Also, partial exposure, drapes, and instruments cause occlusion. Robust pipelines, therefore, need reliable initialization, robustness to outliers/mismatches, and stable tracking under viewpoint changes.

In prior work, we implemented a depth-only, markerless registration system on HoloLens 2 that achieved sub-5\,mm accuracy on rigid 3D-printed anatomy in controlled labs \cite{yang2026easyreg}. That point cloud pipeline performed well on printed models and frozen cadavers (mainly skulls) but assumed large, distinctive, rigid surfaces. In practice, these assumptions are often violated under clinical settings. Here we adapt the system for intraoperative use on smaller, moderately rigid targets: the lower leg, feet, and ear. We introduce a human-in-the-loop initialization (the user roughly aligns a virtual visualization to anatomy) and increase robustness to surface mismatch so alignment remains stable despite moderate skin deformation and occlusions. We then evaluate the system during live fibula free-flap harvest and mandibular reconstruction, covering both donor leg and head/neck sites, and report accuracy and feasibility. To our knowledge, this is among the first clinical evaluations of markerless AR registration on HMDs with real patients, narrowing the gap to clinical readiness while highlighting remaining challenges.

\section{Methods}
\subsection{System overview and pipeline}
\begin{figure}[h]
\centering
\includegraphics[width=\textwidth]{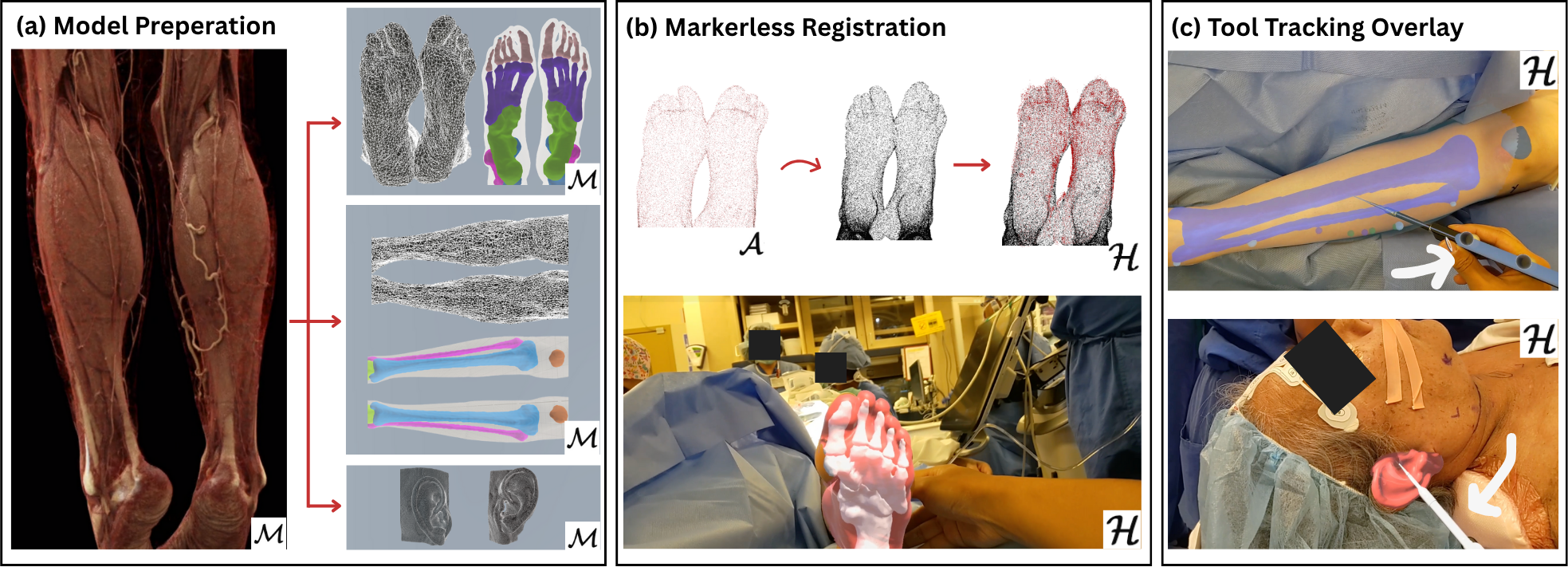}
\caption{Markerless AR registration pipeline. (a) Preoperative CT data is segmented to produce a 3D mesh of the Region of Interest (ROI). Only the skin mesh (feet, lower leg, or ear) is converted to a downsampled point cloud. (b) Intraoperatively, the HoloLens captures a depth image of the exposed anatomy, which is transformed into a point cloud (after correcting sensor bias). The registration module performs a coarse and fine alignment of the preoperative and intraoperative point clouds. (c) We track the handheld tool with fiducials using the HoloLens' depth camera to create a virtual overlay for surface tracing and error measurements, which is evaluated in our experiment. }\label{fig1}
\end{figure}

Our markerless AR registration system comprises three main components: (1) Preoperative model preparation, (2) Intraoperative markerless target registration and tracking (3) Intraoperative tool tracking. An overview of the workflow is shown in \cref{fig1}. Preoperatively, a 3D model of the target anatomy is generated from the patient’s CT scan. This model (only the skin) is converted to a point cloud and downsampled for efficient processing. In the operating room, the operator wearing the HoloLens uses the depth sensor to capture the exposed surface of the target anatomy by simply moving their head. The registration algorithm then computes the rigid transformation that best aligns the preoperative model to the intraoperative depth point cloud. This study was approved by the Stanford IRB (79317).

We define right-handed frames: 1) $\mathcal{M}$: preoperative model frame (CT-derived), 2) $\mathcal{A}$: AHAT depth camera frame, 3) $\mathcal{H}$: HoloLens spatial ``world'' frame, 4) $\mathcal{O}$: external optical tracker frame. A rigid transform from frame $\mathcal{X}$ to $\mathcal{Y}$ is
\begin{equation}
\mathbf{T}_{\mathcal{Y}\mathcal{X}}=
\begin{bmatrix}
\mathbf{R}_{\mathcal{Y}\mathcal{X}} & \mathbf{t}_{\mathcal{Y}\mathcal{X}}\\[2pt]
\mathbf{0}^{\top} & 1
\end{bmatrix}\in SE(3),\quad
\mathbf{R}\in SO(3),\;\mathbf{t}\in\mathbb{R}^{3}.
\end{equation}
Our goal is the pose $\mathbf{T}_{\mathcal{H}\mathcal{M}}=\mathbf{T}_{\mathcal{H}\mathcal{A}}\;\mathbf{T}_{\mathcal{A}\mathcal{M}},$ where $\mathbf{T}_{\mathcal{H}\mathcal{A}}$ is known from device calibration; $\mathbf{T}_{\mathcal{A}\mathcal{M}}$ is estimated by aligning the CT model to the AHAT point cloud through our registration algorithm. Since the depth image of the HoloLens device is pixel-aligned, we obtain scene point cloud and denote as $\mathcal{P}_s=\{\mathbf{p}_i\}_{i=1}^{N_s}\subset\mathbb{R}^3$ in $\mathcal{A}$. The CT-derived surface is sampled as $\mathcal{P}_t=\{\mathbf{q}_j\}_{j=1}^{N_t}\subset\mathbb{R}^3$ in $\mathcal{M}$.

\subsection{Preoperative CT reconstruction and model preparation}
We segment the target external surface (ear, feet, or fibula/lower leg skin) in 3D Slicer \cite{pieper20043d}, export a watertight mesh $\mathcal{M}_{\text{skin}}$, and segment internal structures $\mathcal{M}_{\text{int}}$ (e.g., fibula, tibia, talus) for visualization using TotalSegmenter\cite{wasserthal2023totalsegmentator}. We generate the model point cloud by uniform mesh sampling of the skin and voxel downsampling (voxel size $v$~mm), then farthest-point sampling (FPS) to $N_t$ points:
\begin{equation}
\mathcal{P}_t \leftarrow \mathrm{FPS}\!\big(\mathrm{VoxelDown}(\mathcal{M}_{\text{skin}},v),\,N_t\big),\quad
N_t\approx 5{,}000,\; v\in[1.0,1.5]\;\mathrm{mm}.
\end{equation}
We only converted the external skin surface of the relevant anatomy in each case to a point cloud representation, since our registration is surface-based. In addition to skin, we also segmented key internal structures for visualization purposes (not for registration). For example, in the fibula/lower leg target region, we segmented the fibula bone and the tibia and patella; in the feet target region, we segmented major foot bones (talus, calcaneus, metatarsals, proximal phalanges). 

\subsection{Markerless registration algorithm}
\subsubsection{Problem statement and noise model}
Our registration algorithm takes two inputs: (1) the preoperative target point cloud 
$\mathbf{P}_t$ (e.g., skin surface extracted from CT) and (2) an intraoperative point cloud 
$\mathbf{P}_s$ captured by the HoloLens depth sensor. The algorithm outputs a rigid transformation that aligns 
$\mathbf{P}_t$ to $\mathbf{P}_s$. 

This process involves a coarse global alignment followed by a fine local refinement. 
We denote the points in the target model as 
$\mathbf{P}_t = \{\mathbf{q}_j\}$ 
and the points in the intraoperative scene (patient surface) as 
$\mathbf{P}_s = \{\mathbf{p}_i\}$. Given $\mathcal{P}_s$ and $\mathcal{P}_t$, we estimate
\begin{equation}
\mathbf{T}_{\mathcal{A}\mathcal{M}}^{\star}=
\arg\min_{\mathbf{T}\in SE(3)}
\sum_{(\mathbf{p},\mathbf{q})\in\mathcal{C}}
\rho\!\left(d\big(\mathbf{p},\,\mathbf{T}\mathbf{q}\big)\right),
\end{equation}
where $\mathcal{C}\subset \mathcal{P}_s\times \mathcal{P}_t$ are (noisy) correspondences, $d(\cdot,\cdot)$ is a point-to-plane distance, and $\rho$ is a robust loss (Truncated Least Square in coarse and Turkey's biweight in ICP). We assume bounded noise
\begin{equation}
\mathbf{p}=\mathbf{p}^{\text{true}}+\boldsymbol{\varepsilon}_p,\quad
\mathbf{q}=\mathbf{q}^{\text{true}}+\boldsymbol{\varepsilon}_q,\quad
\|\boldsymbol{\varepsilon}_{(\cdot)}\|\le\sigma,
\end{equation}
with outliers in $\mathcal{C}$. Our registration pipeline includes four phases: (i) region-specific depth bias correction; (ii) human-in-the-loop ROI initialization/cropping; (iii) robust coarse alignment; (iv) fine ICP refinement. Each phase is explained in the following subsections. 

\subsubsection{Depth bias correction}
Before registration, we correct systematic depth errors for the Region of Interest (ROI). As noted in prior studies, the HoloLens depth (AHAT) sensor (high-frequency near-depth sensing) can have depth biases that vary by surface and angle \cite{li2024evd, yang2026easyreg}. Inspired by previous works, we used a tracked stylus to sample ground-truth surface points on the patient’s skin. Accordingly, AHAT depth exhibits local bias $\boldsymbol{\delta}(\mathbf{x})$ over the ROI. With stylus samples $\mathcal{L}=\{\boldsymbol{\ell}_k\}_{k=1}^{n}\subset\mathbb{R}^3$ on patient surface (expressed in $\mathcal{A}$ via precomputed $\mathbf{T}_{\mathcal{A}\mathcal{O}}$) and nearest depth points $\{\mathbf{p}_k\}$, we solve the orthogonal Procrustes
\begin{equation}
\min_{\mathbf{R}\in SO(3),\,\mathbf{t}\in\mathbb{R}^3}
\sum_{k=1}^{n}\left\|\boldsymbol{\ell}_k - (\mathbf{R}\mathbf{p}_k+\mathbf{t})\right\|^2,
\end{equation}
(obtained in closed form via Singular Value Decomposition (SVD) of the centered cross-covariance). We then correct all scene points in the ROI (cropped scene neighborhood around the target) by $\mathbf{p}_i \leftarrow \mathbf{R}\mathbf{p}_i+\mathbf{t},  \
\forall\,\mathbf{p}_i\in\mathcal{P}_s^{\mathrm{ROI}}.$

\begin{figure}[h]
\centering
\includegraphics[width=\textwidth]{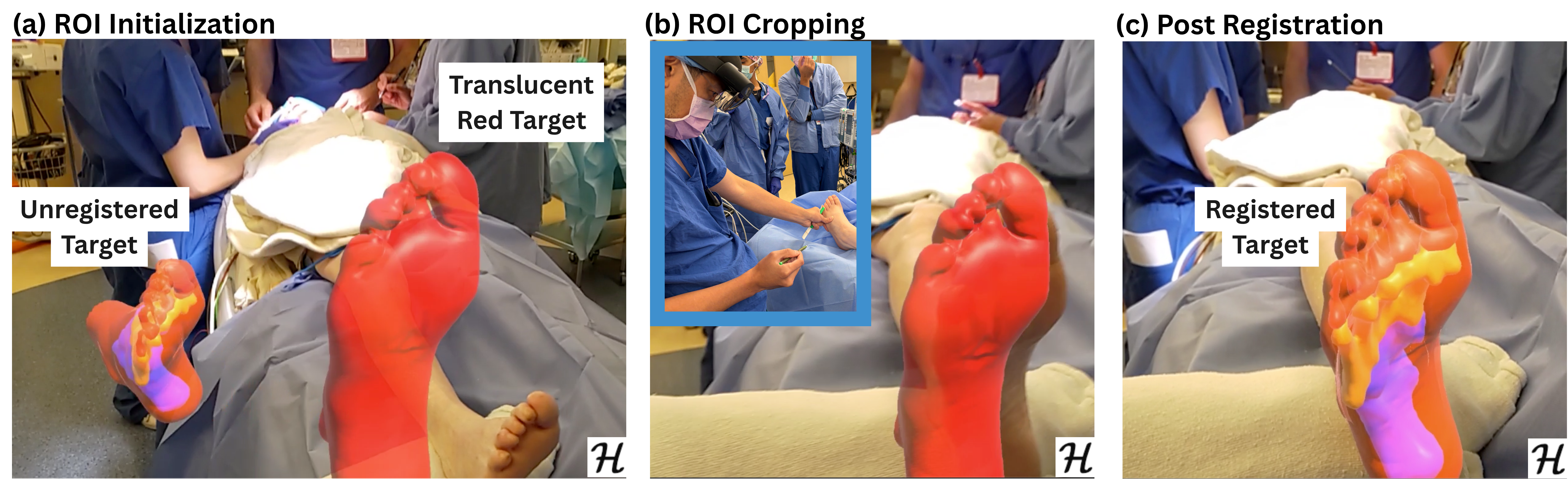}
\caption{Our proposed ROI initialization and cropping process. (a) Mixed reality capture of HoloLens showing both the unregistered target and the translucent reg target when the user first puts on HoloLens. (b) ROI cropping process where the user moves his head to roughly move the translucent red target to the target location. The red target is following head-forward direction constantly. (c) Precise automatic alignment of unregistered target to patient's right foot.}\label{fig2}
\end{figure}

\subsubsection{ROI initialization}
To improve reliability on smaller anatomy, we introduced a user-guided initialization \cref{fig2}. The HoloLens displays a translucent red virtual target model in front of the user before registration. The user moves their head (and thus the HoloLens) so that this virtual target roughly overlaps the real target anatomy, providing $\mathbf{T}_{\mathcal{H}\mathcal{M}}^{(0)}$. \ADD{Users achieve this by moving their viewpoint, which is considered easy and does not require any manual menu interactions.} To express the user-initialized pose in the AHAT frame $\mathcal{A}$,
\begin{equation}
\mathbf{T}_{\mathcal{A}\mathcal{M}}^{(0)}=
\mathbf{T}_{\mathcal{A}\mathcal{H}}\,
\mathbf{T}_{\mathcal{H}\mathcal{M}}^{(0)}
=
\big(\mathbf{T}_{\mathcal{H}\mathcal{A}}\big)^{-1}
\mathbf{T}_{\mathcal{H}\mathcal{M}}^{(0)}.
\end{equation}
We then crop $\mathcal{P}_s$ to a tight ROI centered at $\mathbf{c}=\mathbf{t}_{\mathcal{A}\mathcal{M}}^{(0)}$ with radius $r$ (80--300~mm):
\begin{equation}
\mathcal{P}_s^{\mathrm{ROI}}=\{\mathbf{p}\in\mathcal{P}_s:\ \|\mathbf{p}-\mathbf{c}\|_2\le r\}.
\end{equation} This focuses the algorithm on the relevant scene portion and avoids arbitrary false matches. The human-in-the-loop step provides a bounded search region for the global registration. We found this especially useful for the feet and ear cases, where the anatomy is small and symmetric enough that a fully automatic global alignment might occasionally flip or misplace the model.

\subsubsection{Coarse alignment}
We perform feature-based global registration to achieve coarse alignment despite large misalignments and outliers. Both $\mathbf{P}_t$ and $\mathbf{P}_s$ are downsampled and encoded using Fast Point Feature Histograms (FPFH) \cite{rusu2009fast}. Initial correspondences are found via nearest-neighbor matching in FPFH space. To robustly estimate the transformation, we adopt the TEASER++ algorithm \cite{yang2020teaser}, which solves for scale, rotation, and translation with high tolerance to outliers. \ADD{FPFH balances robustness and speed, whereas learned features remain resource-intensive.}

For two correspondences $(\mathbf{p}_i,\mathbf{q}_j),(\mathbf{p}_k,\mathbf{q}_{\ell})\in\mathcal{C}_0$, we define translation-invariant differences $\Delta\mathbf{p}_{ik}=\mathbf{p}_i-\mathbf{p}_k,\
\Delta\mathbf{q}_{j\ell}=\mathbf{q}_j-\mathbf{q}_{\ell}.$ under a rigid motion $\Delta\mathbf{p}_{ik}\approx \mathbf{R}\,\Delta\mathbf{q}_{j\ell}.$ We then estimate $\mathbf{R}$ via truncated least-squares over edges $\mathcal{E}\subset \mathcal{C}_0\times \mathcal{C}_0$:
\begin{equation}
\mathbf{R}^{\star}=
\arg\min_{\mathbf{R}\in SO(3)}
\sum_{(ik,j\ell)\in\mathcal{E}}
\rho_{\delta_R}\!\left(
\left\|\Delta\mathbf{p}_{ik}-\mathbf{R}\,\Delta\mathbf{q}_{j\ell}\right\|_2
\right),
\end{equation}
solved on $SO(3)$ by graduated-nonconvex optimization with SVD projections.
Given $\mathbf{R}^{\star}$, we recover translation from inliers $\mathcal{I}\subset\mathcal{C}_0$:
\begin{equation}
\mathbf{t}^{\star}=
\arg\min_{\mathbf{t}\in\mathbb{R}^3}
\sum_{(i,j)\in\mathcal{I}}
\rho_{\delta_t}\!\left(\left\|\mathbf{p}_i - (\mathbf{R}^{\star}\mathbf{q}_j+\mathbf{t})\right\|_2\right).
\end{equation}
The output of this stage is a coarse transformation $\mathbf{T}_{\mathcal{A}\mathcal{M}}^{\mathrm{coarse}}= \begin{bmatrix}
\mathbf{R}^{\star} & \mathbf{t}^{\star}\\
\mathbf{0}^{\top} & 1
\end{bmatrix}.$

\subsubsection{Fine alignment}
Starting from the coarse estimate \( \mathbf{T}^{\mathrm{coarse}}_{\mathcal{A}\mathcal{M}} \), we refine alignment with a robust point-to-plane ICP. At iteration \(k\), we form correspondences \( \mathcal{C}_k \subset \mathcal{P}_s \times \mathcal{P}_t \) by matching each scene point \( \mathbf{p}\in\mathcal{P}_s \) to its nearest model point \( \mathbf{q}\in\mathcal{P}_t \). We reject pairs with large point-to-plane residual
\[
r(\mathbf{p},\mathbf{q};\mathbf{R}_k,\mathbf{t}_k)
=\big(\mathbf{R}_k\mathbf{q}+\mathbf{t}_k-\mathbf{p}\big)^\top \mathbf{n}_{\mathbf{p}} ,
\qquad |r|>\tau_k,
\]
where \( \mathbf{n}_{\mathbf{p}} \) is the scene normal at \( \mathbf{p} \) and the threshold \( \tau_k \) decays from \(5\,\mathrm{mm}\) to \(2\,\mathrm{mm}\) as the solution converges. The weighted least-squares update is
\[
(\mathbf{R}_{k+1},\mathbf{t}_{k+1})
=\arg\min_{\mathbf{R}\in SO(3),\,\mathbf{t}\in\mathbb{R}^3}
\sum_{(\mathbf{p},\mathbf{q})\in \mathcal{C}_k}
w_{\mathbf{p}\mathbf{q}}\,
\big[\,\big(\mathbf{R}\mathbf{q}+\mathbf{t}-\mathbf{p}\big)^\top \mathbf{n}_{\mathbf{p}}\,\big]^2,
\]
with robust weights \( w_{\mathbf{p}\mathbf{q}} \). We iterate until the change in the objective is \(<0.5\,\mathrm{mm}\) (or a maximum of 30 iterations). Finally, we compose the incremental update with the coarse pose to obtain the refined transform
\[
\mathbf{T}^{\mathrm{final}}_{\mathcal{A}\mathcal{M}}
\leftarrow
\begin{bmatrix}
\mathbf{R}^\star & \mathbf{t}^\star \\
\mathbf{0}^\top & 1
\end{bmatrix}
\mathbf{T}^{\mathrm{coarse}}_{\mathcal{A}\mathcal{M}},
\]
which we apply to the CT-derived model.

To enable intraoperative tool guidance and to evaluate surface-tracing error, we track the surgical tool using the HoloLens AHAT sensor following \cite{martin2023sttar}. This provides the transformation from the tool to the external optical tracker frame \( \mathcal{O} \). \ADD{This tracking method uses the HoloLens’s ToF depth sensor to detect a pattern of retro-reflective spherical markers attached to a tool, allowing the system to compute the tool’s tip pose with sub-millimeter accuracy.}

\section{Experiments and Results}
\subsection{Preclinical evaluation of surface tracing}
We first evaluated whether surface tracing with an AR tracked tool provides reliable registration error measurements. For each anatomy (lower leg, feet), we formed two paired point cloud sets. In AR, a tracked stylus was used to trace the skin surface of the virtual model (visualized in-air) and then to trace the target internal structure (leg or foot bones). \ADD{In this AR setup, the virtual anatomy was not attached to a physical phantom, so the stylus could ‘penetrate’ the skin to sample points on the internal virtual structure. The stylus tip was visualized in AR to map onto the surface. To ensure high surface coverage, each target surface was traced three times with the tracked stylus; the combined traced points (from all passes) formed a dense point set for comparison with CT ground truth.} For every AR skin surface sample $x$, we computed the shortest Euclidean distance to the traced internal structure, yielding a per-point relative-distance map $d_{\mathrm{AR}}(x)$. Ground truth was obtained from the CT reconstructions by sampling the CT skin surfaces and computing the same shortest distances to the segmented fibula or metatarsals, giving $d_{\mathrm{CT}}(x)$. After rigid alignment (ICP) of the AR and CT surfaces, distances were compared point-wise. Agreement was summarized using the median $|\Delta d|$, interquartile range (IQR), root mean square error (RMSE), the 95th percentile of $|\Delta d|$.

\begin{figure}[h]
\centering
\includegraphics[width=\textwidth]{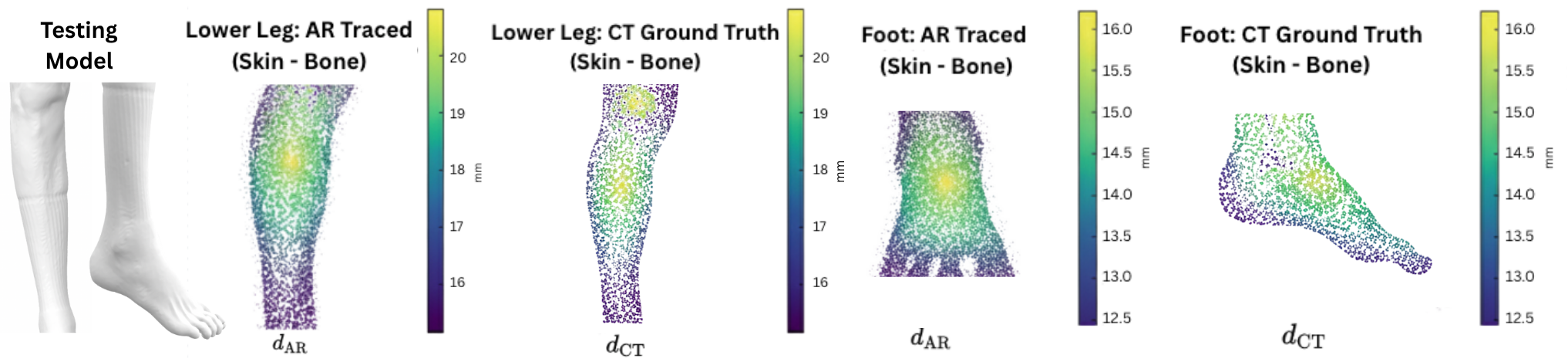}
\caption{Relative distance comparison (AR‑traced vs CT). Left: AR‑traced distance mm) from skin to part of internal structures (e.g. fibula/metatarsals). Right: CT ground‑truth distances.}\label{fig4}
\end{figure}

As shown in \cref{fig4}, for \textit{lower leg:}
$n = 5{,}935$ points, median $|\Delta d| = 0.78\,\mathrm{mm}$ (IQR $0.43\,\mathrm{mm}$), RMSE $0.97\,\mathrm{mm}$. For \textit{feet:} $n = 4{,}057$ points, median $|\Delta d| = 0.80\,\mathrm{mm}$ (IQR $0.55\,\mathrm{mm}$), RMSE $1.20\,\mathrm{mm}$. The small distance differences and tight IQRs between $d_{\mathrm{AR}}$ and $d_{\mathrm{CT}}$ show that stylus-based AR surface tracing reproduces CT-derived relative distances with high fidelity. This supports using a tracked tool in AR as a practical and accurate method for error evaluation of anatomy-aware measurements \textit{in situ}.

\subsection{Clinical evaluation of markerless registration}
HoloLens recordings are presented in supplementary video. We implemented our depth-only, markerless registration pipeline on a head-mounted display (HoloLens~2) coupled to a workstation (Intel\textsuperscript{\textregistered} Core\texttrademark~i9 CPU, NVIDIA\textsuperscript{\textregistered} RTX~4060 GPU). Articulated HAnd
Tracking (AHAT) depth frames streamed at $\sim$30~Hz over a custom TCP protocol; registration ran on the workstation while virtual content rendering occurred on-device in near real-time. Our system setup and AR overlay were deployed immediately after anesthesia induction, during the pre-incision phase, to visualize the anatomy.

We computed registration error as the nearest-neighbor distance between surface-traced points on the virtual overlay and the corresponding points traced on the patient’s skin in the HoloLens world frame. \ADD{Note that the points used for error evaluation (collected via the tracked stylus) were entirely independent from the point clouds used for the registration computation.} Across 7 intraoperative trials (feet1–2, ear1–3, leg1–2), we selected three pools from each trial target. Thus, from each trial’s full traced point set, we formed three disjoint “pools”, yielding $\approx N/3$ points per pool. The combined pooled per-point error had a median of 3.9\,mm and a mean of 4.0\,$\pm$\,1.7\,mm (SD). By anatomy, medians were 3.2\,mm (feet), 4.3\,mm (ear), and 5.3\,mm (leg).

We report several point cloud similarity metrics between the traced point sets:
\begin{itemize}
\item Chamfer Distance (mm$^2$): Mean of squared nearest-neighbor distances
\item Hausdorff Distance (mm): Directed nearest-neighbor distances.
\item Mean/Mean‑Squared Distance.
\item Surface Coverage: Fraction of ground‑truth surface points within a threshold of the overlay surface.
\item Reconstruction Accuracy: Mean distance over covered points
\end{itemize}

\begin{figure}[h]
\centering
\includegraphics[width=\textwidth]{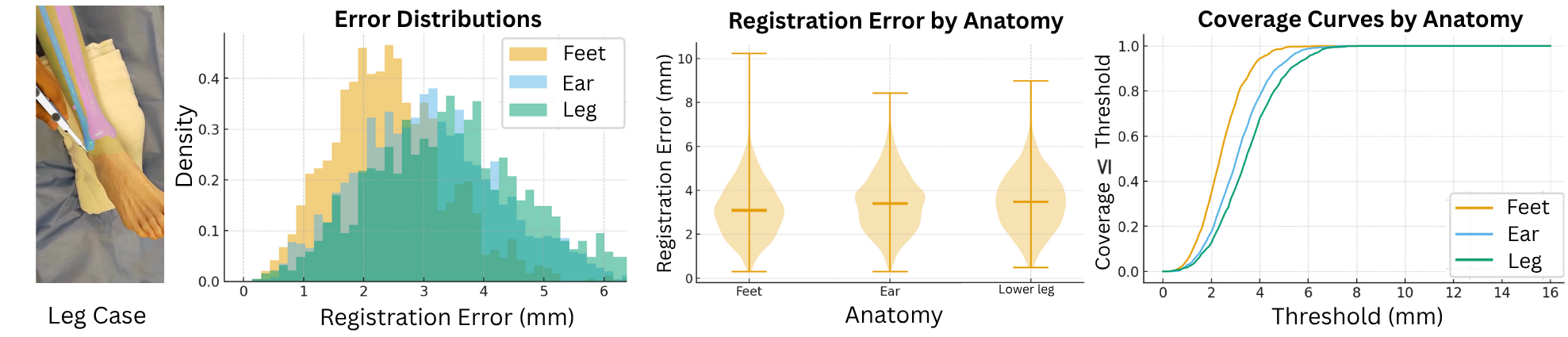}
\caption{Distributional characteristics of registration error across anatomies. From left to right: Overlaid histograms, violin plots showing group variance, and coverage curves showing surface coverage.}\label{fig5}
\end{figure}

We show the distribution of data in \cref{fig5}. We further compared categories using a two-sample permutation test on the difference in medians (balanced subsample \emph{n}=600 per category). Only feet vs. leg showed a statistically significant difference (\(\Delta\)median \(\approx 1.1\) mm, \(\,p<0.001\)). These results indicate that lower‑leg registration is modestly less accurate than feet. \ADD{On average, the registration (including ROI selection) was completed in 5.4 seconds (SD 2.8 s) per trial. Decomposing the error into patient-aligned axes (Right–Left, Anterior–Posterior, Superior–Inferior) revealed slight anisotropy: for laterally viewed targets (e.g., the lower leg), the lateral component of error tended to be around 1 mm larger than the vertical component, suggesting that viewing angle can influence error distribution.}

\begin{table}[h] 
\centering 
\caption{Per-trial surface registration metrics. Distances are in mm except Chamfer (mm$^2$). Values are median, interquartile range (IQR), mean$\pm$SD, and surface coverage within 5\,mm. Range numbers represent result bounds calculated from three pools per trial.} 
\label{tab:trial_metrics} 
\setlength{\tabcolsep}{3pt} 
\renewcommand{\arraystretch}{1.05} 
\begin{tabular*}{\textwidth}{@{\extracolsep{\fill}}lcccccccc@{}} 
\toprule 
Trial & $N$ & Median & IQR & Mean$\pm$SD & Hausdorff & Chamfer & Coverage$_{\le5\mathrm{mm}}$ & Recon. acc. \\ 
\midrule 
feet1 & 600 & 3.1 & 1.8 & 3.1$\pm$1.3 & 30--40 & 22--30 & 92--95\% & 2.6--3.0 \\ 
feet2 & 620 & 3.2 & 1.9 & 3.3$\pm$1.3 & 32--42 & 24--34 & 92--95\% & 2.7--3.1 \\ 
ear1 & 560 & 4.2 & 2.3 & 4.3$\pm$1.7 & 35--50 & 45--60 & 86--90\% & 3.3--3.8 \\ 
ear2 & 540 & 4.4 & 2.4 & 4.5$\pm$1.7 & 38--55 & 50--65 & 84--88\% & 3.5--4.0 \\ 
ear3 & 550 & 4.2 & 2.2 & 4.3$\pm$1.6 & 36--52 & 44--58 & 86--90\% & 3.3--3.8 \\ 
leg1 & 620 & 5.2 & 2.8 & 5.3$\pm$2.0 & 45--60 & 70--90 & 78--86\% & 3.8--4.1 \\ 
leg2 & 600 & 5.4 & 3.0 & 5.6$\pm$2.1 & 48--60 & 75--90 & 72--82\% & 3.9--4.1 \\ 
\bottomrule 
\end{tabular*} 
\end{table}

\section{Discussion and Conclusion}\label{sec:disc}

We tested whether a \emph{depth-only, markerless} pipeline can deliver clinically useful accuracy across multiple anatomies without fiducials. Our key design choices kept the registration consistent across viewpoints within the ROI during surface tracing tasks and avoided any per-case training by (i) a brief human-in-the-loop initialization that bounds the search region, (ii) region-specific AHAT depth bias correction, and (iii) a coarse-to-fine alignment.

\textbf{Accuracy and what it means.} In live surgical settings, we achieved a pooled per-point median error around 3–4\,mm, with high surface coverage within 5\,mm. By anatomy, feet produced the lowest errors, and leg the highest (\cref{tab:trial_metrics,fig5}). Distance-based metrics (Chamfer, Hausdorff) followed the same trend. Using a two-sample permutation test on medians with balanced per-point subsamples, we found feet vs.\ leg significantly different. In practice, these numbers meet the \(\lesssim 5\,\)mm threshold for moderate-risk tasks and enable anatomy visualization without the workflow burden of fiducials. \ADD{While optical marker-based systems can achieve sub-millimeter accuracy at the cost of invasive fiducials, complex setup, and line-of-sight constraints, our method achieves 3–4 mm accuracy without fiducial markers, placing it within the lower error range reported for AR navigation systems \cite{groenenberg2024feasibility, shankar2025ai} while substantially simplifying the clinical workflow.}

\textbf{Why anatomies differ.} Lower leg trials faced smaller exposure windows and more overlying soft tissue to the depth camera, increasing bias and correspondence ambiguity. Feet offered richer curvature and more stable surfaces, making alignment more reliable. Despite rigidness, the ear’s limited surface challenges registration. \ADD{Notably, we used the same registration parameters across all anatomies and achieved successful alignment in each, suggesting that our pipeline is not overly sensitive to hyperparameter settings.}

\textbf{Clinical applicability.} We believe that high efficiency and ergonomic ease of AR are crucial for translation. Surgeons report the lack of loupe function but overall comfort in wearing HMD. Our workflow is fiducial-free but includes a brief stylus calibration to correct depth sensor error on skin. This step takes under one minute and is simpler than placing and tracking physical markers. We performed this calibration for each case to ensure optimal accuracy; however, since AHAT bias on skin is systematic \cite{li2024evd, yang2026easyreg}, a single calibration may generalize to multiple cases if anatomy and conditions are similar.

\textbf{Limitations.} Our pipeline is depth-only and thus sensitive to significant occlusions. \ADD{The stylus-based bias correction improves accuracy but introduces a brief human-in-the-loop step that could be a source of user error or inconsistency.} This initialization step still relies on the user, although it takes seconds and constrains optimization effectively, \ADD{and those more familiar with our system tended to have faster registration times}. Our surface-tracing evaluation can inherit small systematic biases from the tool trajectory and from the AHAT depth itself. \ADD{Robustness under occlusion and poor sensing remains untested. Additionally, since evaluation points were collected in the HMD’s coordinate frame, any head movement could introduce slight latency-related error in their positions; however, users tended to hold their head still during tracing, which likely minimized this effect.} \DEL{Despite rich data, limited cases reduce statistical power and generalizability.}

\textbf{Future work.} We plan to improve our method to account for significant occlusions and soft-tissue deformations. \ADD{Using only depth sensing limits robustness under occlusion; fusing HoloLens RGB with depth data could improve performance.} Future work could also expand evaluation across more anatomies. \DEL{Although our prototype uses a connected workstation to achieve high performance, moving more computation on-device will further simplify the setup and is a goal for future development.} \ADD{We will also pursue a systematic evaluation of the HMD’s usability and ergonomics in clinical settings. Additionally, since our initial analysis indicates errors may be anisotropic with respect to view direction, an evaluation of how the HMD viewing angle correlates with registration error will be valuable in refining the guidance system.}

In conclusion, we demonstrate a markerless AR registration system that generalizes across feet, ear, and lower leg (fibula visualization) in live surgical settings, achieves \(\sim\)3–4\,mm median error.

\section {Declarations}
\subsection{Competing interests}
\ADD{The authors have no related interests.}
\subsection{Ethics \& Consent}
\ADD{The study was approved by the Stanford research ethics committee and followed standard. Informed consent was obtained for each participant.}


\bibliography{sn-bibliography}

\end{document}